\documentclass[conference]{IEEEtran}
\usepackage{amsmath}
\usepackage{xcolor}

\ifCLASSINFOpdf
\else
\fi
\begin{document}
%
\title{Differential Privacy Regularization:\\Protecting Training Data Through\\Loss Function Regularization}

\author{\IEEEauthorblockN{Francisco Aguilera-Martínez}
\IEEEauthorblockA{
Department of Computer Science and Artificial Intelligence\\
ETSIIT, University of Granada\\
faguileramartinez@acm.org}
\and
\IEEEauthorblockN{Fernando Berzal}
\IEEEauthorblockA{
Department of Computer Science and Artificial Intelligence\\
ETSIIT, University of Granada\\
berzal@acm.org}
}


%


\maketitle

\begin{abstract}
Training machine learning models based on neural networks requires large datasets, which may contain sensitive information. The models, however, should not expose private information from these datasets. Differentially private SGD [DP-SGD] requires the modification of the standard stochastic gradient descent [SGD] algorithm for training new models. In this short paper, a novel regularization strategy is proposed to achieve the same goal in a more efficient manner.
\end{abstract}

\section{Introduction}

We have recently witnessed the widespread adoption of deep learning models across many different applications. These models have been particularly successful in an increasing number of natural language processing (NLP) tasks \cite{dl-security-fu2021}, including text summarization, machine translation, and language generation. Large language models (LLMs) have gained significant attention for their exceptional capabilities to generate and interpret text as humans do \cite{llm-security-wu2024}. While LLMs offer significant advantages, they are not without flaws and remain vulnerable to security and privacy attacks \cite{llm-security-das2024}. 

Training and fine-tuning LLMs requires massive quantities of data sourced from the internet and proprietary data sources, as well as carefully annotated text data to enhance their performance for specific tasks. This reliance on extensive datasets can increase their susceptibility to security and privacy vulnerabilities. Despite their widespread use, the vulnerabilities of LLMs have not been extensively explored on a large scale. 

Malicious actors can exploit deep learning models to extract sensitive information that is used during their training and is, in some sense, memorized by them. One of the most prominent and dangerous attack methods is called gradient leakage \cite{llm-security-das2024}. This attack attempts to infer whether a specific data instance was part of the training data used to train (or fine-tune) the model under attack. To mitigate the effectiveness of these attacks, model developers sometimes employ differential privacy \cite{dp-sgd-utility} during the training process as a protective measure. 

We investigate how to preserve privacy by implementing differential privacy regularization for deep learning models, including LLMs. Our approach is inspired by the differentially private stochastic gradient descent (DP-SGD) algorithm \cite{dp-sgd}. DP-SGD introduces random noise in the gradients as a defense mechanism during model training. Our work started from the observation that the introduction of Gaussian noise in the gradients, as the DP-SGD algorithm does, is not completely effective in mitigating gradient leakage (GL) attacks. 

In this work, we propose a new method that intends to achieve differential privacy through the regularization of the loss (or error) function used to train artificial neural networks. In order to offer protection against gradient leakage (GL) attacks, our regularization term depends directly on both the network parameters and its inputs. Our proposal, called PDP-SGD, is somehow equivalent to the introduction of Gaussian noise that is proportional to the magnitude of each parameter in the model. However, the explicit introduction of noise is not needed, actually, and its computational cost can be avoided by PDP-SGD. 

\section{Background}    

Differential privacy (DP) has emerged as a fundamental technique for safeguarding sensitive information in machine learning models, particularly in scenarios involving large language models. Differential privacy limits the information that is leaked about specific individuals. In terms of machine learning models, we try to ensure that an individual data does not significantly affect the outcome of a computation, therefore providing guarantees of privacy while allowing useful insights to be derived from aggregate data. In short, we are focused on the study of how differential privacy can be used to protect sensitive information in training data.

In the context of deep learning, implementing differential privacy poses unique challenges due to the complexity of the models and the sensitivity of the data often used to train deep learning models. Below, we summarize some key works that have shaped the field of differential privacy in deep learning.

\subsection{Differential Privacy in Deep Neural Networks}

Abadi et al. \cite{dp-sgd} explored the integration of differential privacy in deep neural networks. 

The core idea of differential privacy is to ensure that the inclusion or exclusion of a single training example does not significantly influence the outcome of a model. Formally, an algorithm \( A \) is said to satisfy \((\epsilon, \delta)\)-differential privacy if, for any two adjacent datasets \( D \) and \( D' \), and for any subset of possible results \( S \), the following holds:
\[
\Pr[A(D) \in S] \leq e^{\epsilon} \Pr[A(D') \in S] + \delta
\]
Here, \( \epsilon \) represents the privacy budget, which quantifies the level of privacy protection, and \( \delta \) accounts for a small probability of failure to maintain privacy. 

In practice, implementing differential privacy in deep neural networks involves techniques like gradient clipping and Gaussian noise addition. The resulting algorithm is called DP-SGD, differentially-private stochastic gradient descent. Gradient clipping limits the gradient norm to a predefined threshold \( C \), reducing sensitivity, while Gaussian noise is added to the gradients to obscure individual data contributions. This approach, however, introduces a trade-off, as higher privacy levels tend to decrease model accuracy due to the increased noise that is introduced during model training.

\subsection{Differential Privacy in LLMs: The EW-Tune Framework}

Differential privacy has also been applied to large language models (LLMs), which present additional challenges due to their size and complexity. Behnia et al. \cite{dp-ewtune} proposed the EW-Tune framework to implement DP in LLMs. They introduced the Edgeworth Accountant, a method for calculating precise privacy guarantees in the context of finite samples. By leveraging the Edgeworth expansion, the authors provide non-asymptotic guarantees, improving upon traditional approaches that often rely on asymptotic bounds.

The DP-SGD algorithm is used for applying DP in LLMs, by introducing Gaussian noise into the gradients during model training. The Edgeworth Accountant just refines the noise addition process, calculating the necessary amount of noise based on the given privacy budget. This method balances the trade-off between noise and model utility more effectively, allowing for reduced noise and, consequently, better model performance without compromising privacy guarantees.

\subsection{User-Level Differential Privacy in LLMs}

Building on previous work, Charles et al. \cite{dp-user-level} examine DP in the context of LLM training by introducing two sampling approaches: example-level sampling (ELS) and user-level sampling (ULS). These methods aim to protect user-level privacy, ensuring that the contribution of individual users to the model is protected. ELS involves clipping gradients at the example level, while ULS operates at the user level, allowing for gradient aggregation over all examples provided by a single user.

The authors introduce a novel user-level DP accountant for ELS that leverages a divergence measure known as the hockey-stick divergence. This measure enables the derivation of precise privacy guarantees for ELS. Comparisons between ELS and ULS show that, under fixed computational budgets, ULS tends to provide stronger privacy guarantees and better model performance, particularly when stringent privacy protection is required or when larger computational resources are available.

\subsection{Differential Privacy through Classic Regularization}

While DP-SGD is a robust method for ensuring privacy, it often leads to significant performance degradation due to noise used during training. Lomurno et al. \cite{dp-sgd-utility} compared DP techniques and traditional regularization methods, such as dropout and L2 regularization. According to their study, classic regularization, commonly used to prevent overfitting, provides similar levels of protection against membership inference and model inversion attacks, which are key privacy concerns in machine learning.

Their empirical results suggest that regularization methods may offer a more effective trade-off between privacy and model performance in certain scenarios. Unlike DP-SGD, which incurs in high computational costs and significant accuracy loss, regularization techniques provide privacy protection with a minimal impact on performance. As such, these methods may be more suitable in contexts where model performance and training efficiency are critical.

\section{A New Perspective on the Differentially Private SGD Algorithm}

DP-SGD \cite{dp-sgd} offers one way to control the influence of the training data during the training process: At each step of the SGD, we compute the gradient $\nabla_\theta \mathcal{L}(\theta)$ for a random subset of examples, clip the $\ell_2$ norm
of each gradient, compute the average, add noise in order to
protect privacy, and take a step in the opposite direction of
this average noisy gradient:
$$ \epsilon(t) \sim \mathcal{N}(0,\sigma^2)$$
$$ \tilde{g}(t) = g(t) + \epsilon(t)$$
where $g(t)$ is the gradient for the current batch of training examples and $\epsilon(t)$ is the Gaussian noise we introduce.

The resulting weight update is, therefore:
$$ \Delta \theta(t) = - \eta_t \tilde{g}(t) $$
i.e.
$$ \theta(t+1) = \theta(t) - \eta_t \tilde{g}(t)$$
where $\eta_t$ is the current learning rate.


Let us now assume a linear neuron (or a nonlinear neuron operating within its linear regime):
$$ y = \theta \cdot x = \sum_{i=0}^n \theta_i x_i$$
Before the weight update:
$$ y(t) = \theta(t) \cdot x $$
With noise in the gradients, the output after the weight update is
\begin{align*}
\tilde{y}(t+1) &= \theta(t+1) \cdot x \\
               &= (\theta(t) - \eta_t \tilde{g}(t)) \cdot x \\
               &= (\theta(t) - \eta_t (g(t) + \epsilon(t)) \cdot x \\
               &= (\theta(t) - \eta_t (g(t)) \cdot x  - \eta_t \epsilon(t) \cdot x \\
               &= y(t+1) - \eta_t \epsilon(t) \cdot x 
\end{align*}

Simplifying our notation, we have:
\begin{align*}
\tilde{y} &= (\theta - \eta \tilde{g}) \cdot x \\
          &= y - \eta \epsilon x
\end{align*}

Using a quadratic error function $\mathcal{L} = (y-t)^2$ for the training algorithm using gradient noise:
\begin{align*}
E[(\tilde{y}-t)^2] &= E[(\tilde{y}-t)^2] \\
             &= E[((y - \eta \epsilon x) - t)^2] \\
             &= E[((y - t) - \eta \epsilon x)^2] \\
             &= E[(y - t)^2] - E[2(y-t) \eta \epsilon x] + E[(\eta \epsilon x)^2]
\end{align*}

Let us recall that the noise $\epsilon$ is sampled from a Gaussian with mean 0 and variance $\sigma^2$. Hence:
$$ E[2(y-t) \eta \epsilon x] = 2(y-t) \eta E[\epsilon] x = 0$$

Therefore
$$ E[(\tilde{y}-t)^2] = E[(y - t)^2] + E[(\eta \epsilon x)^2] $$ 

The first term is just the traditional quadratic error function $\mathcal{L} = (y - t)^2$, whereas the second term can be interpreted as an $L^2$ regularization term for the input:
\begin{align*}
E[( & \eta \epsilon x)^2] \\
                       &= \eta^2 E[(\epsilon x)^2] \\
                       &= \eta^2 E \left[ \left( \sum_i \epsilon_i x_i \right)^2 \right] \\
                       &= \eta^2 E \left[ \sum_i \epsilon_i^2 x_i^2 +  2 \sum_{i<j} \epsilon_i x_i \epsilon_j x_j \right] \\
                       &= \eta^2 \left( \sum_i E\left[\epsilon_i^2 x_i^2 \right] + 2 \sum_{i<j} E\left[ \epsilon_i x_i \epsilon_j x_j \right] \right) \\
                       &= \eta^2 \left( \sum_i E[\epsilon_i^2] E[x_i^2] + 2 \sum_{i<j} E[\epsilon_i] E[x_i] E[\epsilon_j] E[x_j]  \right) \\
                       &= \eta^2 \left( \sum_i \sigma_i^2 x_i^2 + 2 \sum_{i<j} 0 x_i 0 x_j \right) \\
                       &= \eta^2 \sum_i \sigma_i^2 x_i^2
\end{align*}

The error function given the noisy gradients in DP-SGD is finally
$$ E[(\tilde{y}-t)^2] = (y - t)^2 + \eta^2 \sum_i \sigma_i^2 x_i^2 $$

If we assume that the gradient noise variance is the same for all the inputs:
$$ E[(\tilde{y}-t)^2] = (y - t)^2 + \eta^2 \sigma^2 \sum_i x_i^2 $$
$$ \mathcal{L}_\text{noisy gradient} = \kappa \sum_i x_i^2$$

$$ \mathcal{L}_\text{DP} = \mathcal{L} + \mathcal{L}_\text{noisy gradient}$$
Please, compare the above expression with the standard $L^2$ regularization strategy:
$$ \mathcal{L}_\text{$L^2$ regularization} = \mathcal{L} + \lambda \sum_i \theta_i^2$$

Of course, both regularization terms can be easily combined:
$$ \mathcal{L}_\text{DP+$L^2$ regularization} = \mathcal{L} + \lambda \sum_i \theta_i^2 + \kappa \sum_i x_i^2 $$

As training with input noise is equivalent to weight decay, also known as Tikhonov or $L^2$ regulatization \cite{noise-regularization}, training with noisy gradients is somehow equivalent to performing Tikhonov regularization on the input.

However, it should be noted that the DP regularization term is independent from the network parameters $\theta$. Therefore, its gradient with respect to the network parameters is zero, i.e. $\nabla_\theta \mathcal{L}_\text{noisy gradient} = 0$. Hence, the resulting optimization algorithm is exactly the same for the standard stochastic descent algorithm (SGD) and for its gradient noise variant (DP-SGD). In other words, we are just introducing some artificial noise in the training algorithm, which adds to the noisy estimate of the gradient computed by the stochastic gradient descent algorithm.

The above discussion might explain why some researchers have found that, even though DP-SGD ``is a popular mechanism for training machine learning models with bounded leakage about the presence of specific points in the training data[,] [t]he cost of differential privacy is a reduction in the model's accuracy'' \cite{dp-impact-on-accuracy}. Moreover, ``[a]ccording to the literature, [DP-SGD] has proven to be a successful defence against several models’ privacy attacks, but its downside is a substantial degradation of the models’ performance... and [researchers] empirically demonstrate the often superior privacy-preserving properties of dropout and l2-regularization'' \cite{dp-sgd-utility}.


\section{Differentially Private Regularization}

In the previous section, we observed that the addition of Gaussian noise to the gradients in DP-SGD is not really effective, since it just introduces an additional noise to the noisy gradient estimate of the conventional SGD, without really changing the loss function we are implicitly optimizing.

In this Section, we propose the introduction of noise proportional to each parameter value, so that the resulting algorithm is not equivalent to SGD in its linear regime. 

Our proportional differentially private PDP-SGD algorithm\footnote{Our PDP acronym is intentional, in honor to the Stanford Parallel Distributed Processing (PDP) lab led by Jay McClelland,  who is known for his work on statistical learning, applying connectionist models (i.e. neural networks) to explain cognitive phenomena such as spoken word recognition and visual word recognition, and the books he edited in the 1980s, which spurred the scientific interest in connectionism \cite{pdp1}\cite{pdp2}.} starts by introducing Gaussian noise as follows:
$$ \epsilon_i(t) \sim \mathcal{N}(0,(\theta_i \sigma)^2)$$
$$ \tilde{g}_i(t) = g_i(t) + \epsilon_i(t)$$

For each network parameter, $\theta_i$, we add Gaussian noise whose standard deviation is proportional to the parameter value (i.e. larger parameters receive larger noise). The gradient noise variance is, therefore, $\sigma_i^2 = (\theta_i \sigma)^2$

By definition, in Gaussian noise, the values are identically distributed and statistically independent (and hence uncorrelated), so $E[\epsilon_i \epsilon_j] = E[\epsilon_i] E[\epsilon_j]$.

Using the same reasoning we followed in the previous Section, we have:
\begin{align*}
\tilde{y} &= y - \eta \epsilon x
\end{align*}
and
\begin{align*}
E[(\tilde{y}-t)^2] &= 
  E[(y - t)^2] - E[2(y-t) \eta \epsilon x] + E[(\eta \epsilon x)^2]
\end{align*}

Now, the noise $\epsilon_i$ for each gradient is sampled from a Gaussian with mean 0 and variance $\sigma_i^2$, which is different for each parameter. Hence:
\begin{align*}
E[2(y-t) \eta \epsilon x] 
  &= 2(y-t) \eta E[\epsilon x] \\
  &= 2(y-t) \eta E \left[\sum \epsilon_i x_i \right] \\
  &= 2(y-t) \eta \sum E[\epsilon_i x_i] \\
  &= 2(y-t) \eta \sum E[\epsilon_i] E[x_i]\\
  &= 2(y-t) \eta \sum 0 E[x_i]\\
  &= 0
\end{align*}
\begin{align*}
E[(& \eta \epsilon x)^2] \\
  &= \eta^2 E[(\epsilon x)^2] \\
  &= \eta^2 E \left[ \left( \sum_i \epsilon_i x_i \right)^2 \right] \\
  &= \eta^2 E \left[ \sum_i \epsilon_i^2 x_i^2 +  2 \sum_{i<j} \epsilon_i x_i \epsilon_j x_j \right] \\
  &= \eta^2 \left( \sum_i E\left[\epsilon_i^2 x_i^2 \right] + 2 \sum_{i<j} E\left[ \epsilon_i x_i \epsilon_j x_j \right] \right) \\
  &= \eta^2 \left( \sum_i E[\epsilon_i^2] E[x_i^2] + 2 \sum_{i<j} E[\epsilon_i] E[x_i] E[\epsilon_j] E[x_j]  \right) \\
  &= \eta^2 \left( \sum_i \sigma_i^2 x_i^2 + 2 \sum_{i<j} 0 x_i 0 x_j \right) \\
  &= \eta^2 \sum_i \sigma_i^2 x_i^2 \\
  &= \eta^2 \sum_i (\theta_i \sigma)^2 x_i^2 \\
  &= \eta^2 \sigma^2 \sum_i \theta_i^2 x_i^2 \\
\end{align*}

The error function given the proportional noisy gradients in PDP-SGD is, therefore,
$$ E[(\tilde{y}-t)^2] = (y - t)^2 + \eta^2 \sigma^2 \sum_i \theta_i^2 x_i^2 $$

If we define a proportional differentially private regularization term as follows
$$ \mathcal{L}_\text{proportional noisy gradient} = \kappa \sum_i \theta_i^2 x_i^2$$
then we have
$$ \mathcal{L}_\text{PDP} = \mathcal{L} + \mathcal{L}_\text{proportional noisy gradient}$$

Now, our regularization term depends on the network parameters, so its gradient is not zero:
$$ \nabla_{\theta_i} \mathcal{L}_\text{proportional noisy gradient} = 2 \kappa x_i^2 \theta_i$$
$$ \nabla_{\theta_i} \mathcal{L}_\text{PDP regularization} 
 = \nabla_{\theta_i} \mathcal{L} + 2 \kappa x_i^2 \theta_i $$

Let us finally observe that this term is still different from the standard $L^2$-regularization, which does not depend on the inputs:
$$ \mathcal{L}_{L^2} = \lambda \sum_i \theta_i^2$$
$$ \nabla_{\theta_i} \mathcal{L}_{L^2} = 2 \lambda \theta_i$$

In fact, both regularization techniques can be easily combined and incorporated into the standard training procedure for deep neural networks:

$$ \mathcal{L}_\text{PDP+$L^2$ regularization} = \mathcal{L} + \lambda \sum_i \theta_i^2 + \kappa \sum_i \theta_i^2 x_i^2 $$
\begin{align*}
\nabla_{\theta_i} \mathcal{L}_\text{PDP+$L^2$ regularization} 
&= \nabla_{\theta_i} \mathcal{L} + 2 \lambda \theta_i + 2 \kappa x_i^2 \theta_i \\
&= \nabla_{\theta_i} \mathcal{L} + 2 (\lambda + \kappa x_i^2 ) \theta_i \\
\end{align*}

Given that the goal of differential privacy is protecting training data (i.e. the inputs $x_i$), we hypothesize that the proposed proportional differentially-private regularization term, PDP, should be more effective than the popular DP-SGD algorithm in practice. In addition, it would also be more efficient, since the introduction of noise when computing the gradients in SGD is replaced by an extra regularization term in the loss function used to train the network, which can then be trained using the standard SGD optimization algorithm of our choice (e.g. Adam).



\bibliographystyle{ieeetr}
\bibliography{bibliography}

\appendix

\subsection{Expectations (and variances)}

\subsubsection{Linearity of expectations}

The expected value operator (or ''expectation operator'') $\operatorname{E}[\cdot]$ is linear in the sense that, for any random variables $X$ and $Y$, and a constant $a$:
$$\operatorname{E}[X + Y] = \operatorname{E}[X] + \operatorname{E}[Y]$$
$$\operatorname{E}[aX] = a \operatorname{E}[X]$$
This means that the expected value of the sum of any finite number of random variables is the sum of the expected values of the individual random variables, and the expected value scales linearly with a multiplicative constant. 

The variance of a random variable  $X$  is the expected value of the squared deviation from the mean of $X$: $Var[X] = E[(X-E[X])^2] = E[X^2] - E[X]^2$. Therefore, $ E[X^2] = Var[X] - E[X]^2$.

Variance is invariant with respect to changes in a location parameter, i.e. $Var[X+a]=Var[X]$. However, when all values are scaled by a constant, the variance is scaled by the square of that constant: $Var[aX]=a^2 Var[X]$. In general, the variance of the sum of two random variables is
$$Var[X+Y] = Var[X] + Var[Y] + 2Cov[X,Y]$$
where $Cov[X,Y]$ is the covariance $Cov[X,Y] = E[(X-E[X])(Y-E[Y])] = E[XY] - E[X]E[Y]$ using the linearity property of expectations.

\subsubsection{Non-multiplicativity of expectations}

If $X$ and $Y$ are independent, then $E[XY] = E[X] \cdot E[Y]$. However, in general, the expected value is not multiplicative, i.e. $E[XY]$ is not necessarily equal to $E[X] \cdot E[Y]$.  In fact, $Cov[X,Y] = E[XY] - E[X]E[Y]$.

The variance of the product of two independent random variables is $Var[XY]= E[(XY)^2] - E[XY]^2 = E[X^2]E[Y^2] - (E[X]E[Y])^2 = Var[X]Var[Y] + Var[X] E[Y]^2 + Var[Y] E[X]^2$, which can be rewritten as $ Var[X] E[Y]^2 + Var[Y] E[X]^2 + (Cov[X,Y] / \rho[X,Y])^2$, where $\rho[X,Y]$ is the Pearson correlation coefficient, $\rho[X,Y]=Cov[X,Y]/\sqrt{Var[X]Var[Y]}$.

\subsection{Normal distributions}

A normal or Gaussian distribution with mean $\mu$ and variance $\sigma^2$ is a continuous probability distribution for a real-valued random variable whose probability density function is
$$ P(x) = \frac{1}{\sigma \sqrt{2\pi}} e^{\frac{-1}{2} \left( \frac{x-\mu}{\sigma} \right)^2} $$

If $X$ is distributed normally with mean $\mu$ and variance $\sigma^2$, then
$aX+b$, for any real numbers $a$ and $b$, is also normally distributed, with mean $a\mu+b$ and variance $a^2\sigma^2$. That is, the family of normal distributions is closed under linear transformations. Hence, 
$E[kX] = k\mu$ because $E[X] = \mu$.

\subsubsection{Product of normal distributions}

The distribution of a product of two normally distributed random variables $X$ and $Y$ with zero means and variances $\sigma_x^2$ and $\sigma_y^2$ is given by

\begin{align*}
P_{XY}(u) 
  &= \int_{-\infty}^{\infty} \int_{-\infty}^{\infty} 
  P_X(x) P_Y(y) \delta(xy-u) dx dy	\\
  &= \frac{1}{\pi \sigma_x \sigma_y} K_0 \left( \frac{|u|}{\sigma_x\sigma_y} \right)	
\end{align*}
where $\delta(x)$ is Dirac's delta function and $K_n(z)$ is a modified Bessel function of the second kind.
\begin{align*}
K_0(z) &= \int_0^\infty \cos(z \sinh t) dt \\
       &= \int_0^\infty \frac{\cos(zt)}{\sqrt{t^2+1}} dt
\end{align*}

\subsubsection{Square of normal distributions}
For a general normal distribution $X \sim \mathcal{N}(\mu, \sigma^2)$, you can use the fact that $X = \mu + \sigma N$ where $N$ is a standard normal (zero mean, unit variance) to get
$$ X^2 = \mu^2 + 2\sigma\mu N + \sigma^2 N^2$$
For a zero-mean normal distribution, $X \sim \mathcal{N}(0, \sigma^2)$, $ X^2 = \sigma^2 N^2$. $X^2/\sigma^2$ follows a Chi-squared distribution with 1 degree of freedom, i.e. $ X^2 / \sigma^2 \sim \chi_1^2 $ (a non-central Chi-squared distribution in general, when the mean is not zero).

$$ X \sim \mathcal{N}(0, \sigma^2) $$
$$ X^2 \sim \sigma^2 \chi_1^2 $$
Since $\mu = E[X] = 0$ and $Var[X]=E[X^2]-E[X]^2$,
$$E[X^2] = Var[X] = \sigma^2 $$
Finally,
$$Var[X^2] = E[X^4] - E[X^2]^2$$
$$E[X^4] = 3\sigma^4$$
$$Var[X^2] = E[X^4] - E[X^2]^2 = 3\sigma^4 - \sigma^4 = 2\sigma^4$$
NOTE: $X^2 \sim \sigma^2 \chi_1^2$. Since $E[\chi_1^2]=1$ and $Var[\chi_1^2]=2$, then $E[X^2]=\sigma^2$ and $Var[X^2]=2\sigma^4$.

\end{document}